\documentclass[sigconf]{acmart}

\usepackage{booktabs} %

\usepackage{hhline}
\usepackage{times}
\usepackage{amssymb}
\usepackage{amsmath}
\usepackage{pbox}
\usepackage{multirow}
\usepackage{bbm}
\usepackage{multirow,bigdelim}
\usepackage{pgfplots}
\usepackage{graphicx} %
\usepackage{caption}
\usepackage{subcaption}
\usepackage{algorithm}
\usepackage{algorithmic}
\usepackage{enumitem}
\usepackage{makecell}

\DeclareMathOperator*{\argmin}{arg\,min}
\DeclareMathOperator*{\argmax}{arg\,max}

\newcommand*{\origrightarrow}{}

\renewcommand*{\textrightarrow}{\fontfamily{cmr}\selectfont\origrightarrow}

\copyrightyear{2017} 
\acmYear{2017} 
\setcopyright{rightsretained} 
\acmConference{CIKM'17 }{November 6--10, 2017}{Singapore, Singapore}\acmDOI{10.1145/3132847.3132959}
\acmISBN{978-1-4503-4918-5/17/11}

\begin{document}
\title{Learning Edge Representations via Low-Rank Asymmetric Projections}

\author{Sami Abu-El-Haija}
\affiliation{%
  \institution{Google Research}
  \city{Mountain View} 
  \state{California}
}
\email{haija@google.com}

\author{Bryan Perozzi}
\affiliation{%
  \institution{Google Research}
  \city{New York City} 
  \state{New York} 
}
\email{bperozzi@acm.org}

\author{Rami Al-Rfou}
\affiliation{%
  \institution{Google Research}
  \city{Mountain View} 
  \state{California} 
}
\email{rmyeid@google.com}

\begin{abstract}
We propose a new method for embedding graphs while preserving directed edge information.
Learning such continuous-space vector representations (or embeddings) of nodes in a graph is an important first step for using network information (from social networks, user-item graphs, knowledge bases, etc.) in many machine learning tasks.

Unlike previous work, we (1) explicitly model an edge as
a function of node embeddings, and we (2) propose a novel objective,
the \emph{graph likelihood}, which contrasts information from sampled random walks with non-existent edges.
Individually, both of these contributions improve the learned
representations, especially when there are memory constraints on
the total size of the embeddings. 
When combined, our contributions
enable us to significantly improve the state-of-the-art by learning
more concise representations that better preserve the graph structure.

We evaluate our method on a variety of link-prediction task including
social networks, collaboration networks, and protein interactions,
showing that our proposed method learn representations with error
reductions of up to 76\% and 55\%, on directed and undirected
graphs. 
In addition, we show that the representations learned
by our method are quite space efficient, producing embeddings which have higher structure-preserving accuracy but are 10 times smaller.

\end{abstract}


\maketitle

\section{Introduction}
Recent advancements in learning embedding vectors for words have resulted in a proliferation of methods which learn continuous space representations of graphs (e.g. DeepWalk \cite{deepwalk}).
These approaches process a graph and encode each node as a (real-valued) embedding vector,
enabling easy integration with existing machine learning algorithms. 

Such embedding methods learn a vector space that highly preserves the graph structure.
Two nodes would have large similarity in the embedding space
(or small distance) if they are \textit{strongly} connected in the original (discrete) graph. Edges can be weighted or unweighted.
Traditional \emph{eigen} methods \cite{vlsi-spectral, shi-malik, eigenmaps} learn embeddings that minimize the
euclidean distance of connected nodes, which can be solved (with orthonormal constraints) by eigen-decomposition of
the symmetric graph Laplacian.
Recent \emph{random-walk} embedding methods \cite{deepwalk,node2vec}  learn representations which encode the
random walk transition matrix. These methods embed two nodes close if they co-occur frequently in short random
walks.
In general, \emph{random-walk} methods outperform ``eigen'' methods on producing vector representations that preserve the graph structure.

\begin{figure}[t!]
	\begin{center}
		\centerline{\includegraphics[width=\columnwidth]{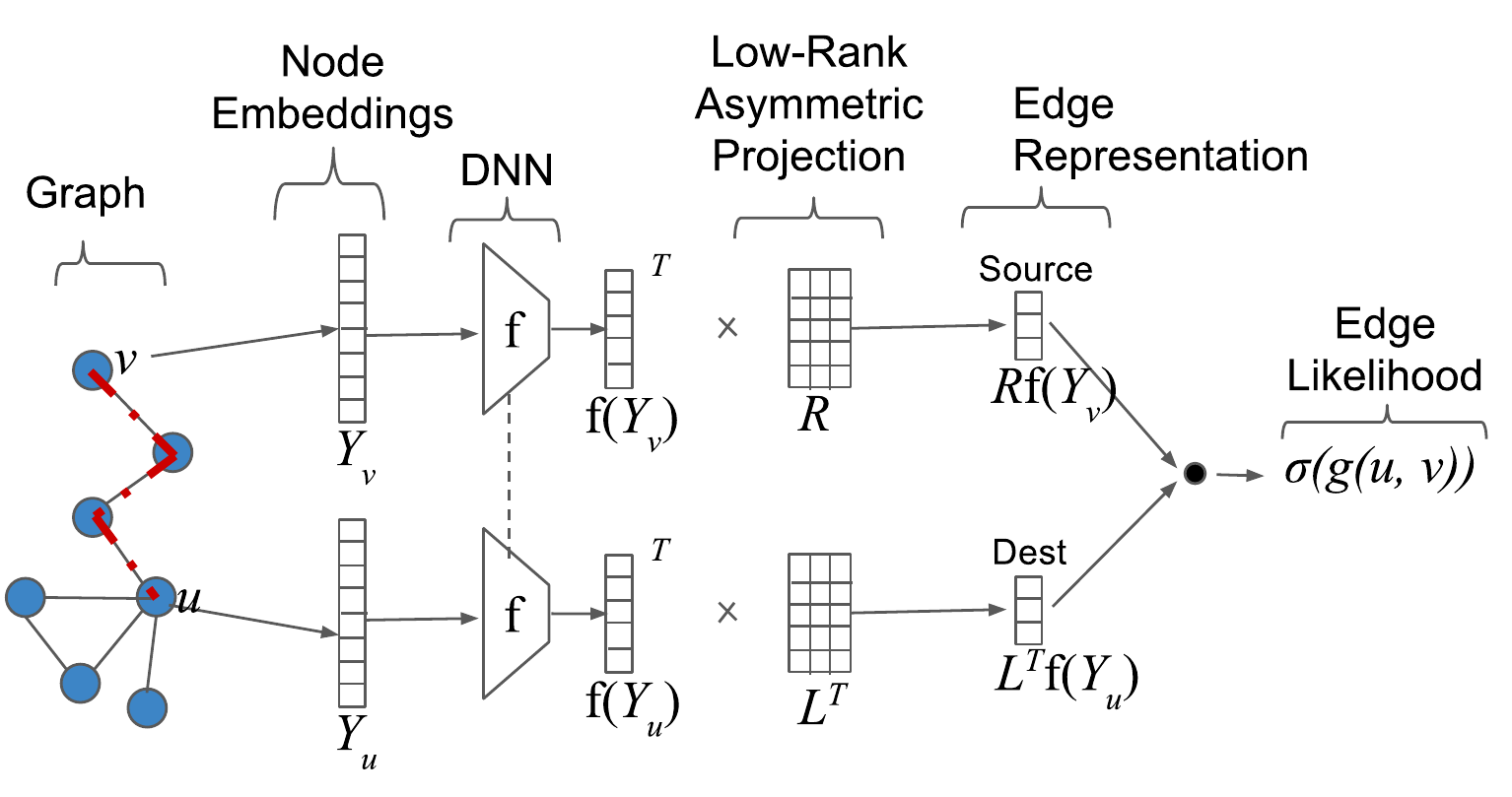}}
		\caption{
			Depiction of our method. On the left: a graph, showing a random walk in dotted-red, where nodes $u, v$ are ``close'' in the walk (i.e. within a configurable
			                       context window parameter).
			               We access the trainable embeddings $Y_u$ and $Y_v$ for the
			                       nodes and feed them as input to Deep Neural Network (DNN) $f$.
			               The DNN outputs manifold coordinates $f(Y_u)$ and $f(Y_v)$ for nodes $u$ and $v$, respectively.
			               A low-rank asymmetric projection transforms $f(Y_u)$ and $f(Y_v)$ to their source and destination representations,
			which are used by $g$ to represent an edge.}
		\label{fig:model}
	\end{center}
	\vskip -0.2in
\end{figure}

However, recent random-walk embedding methods have two shortcomings.
First, these methods do not explicitly model edges.
This \emph{node-centric} assumption represents an edge $(u, v)$ identically to reverse counterpart $(v, u)$, and is unable to capture asymmetric relationships.
Second, to preserve the graph structure they embed nodes into a relatively high-dimensional space,
sometimes producing an embedding dictionary larger than the sparse adjacency matrix.

In this work we propose to address these limitations by explicitly modeling edges in the network as a function of the nodes.
Specifically, we model edges by
(i) using a Deep Neural Network (DNN) to map nodes onto a low-dimensional manifold,
(ii) defining an edge function between two nodes as a projection in the manifold coordinates,
and (iii) jointly-optimizing the edge function and the manifold by maximizing a new objective we propose, the \textit{graph likelihood},
which we define as a product of the edge function over all node pairs.

More formally, we learn an embedding vector $Y_u \in \mathbb{R}^D$ for every graph node $u$,
a manifold-mapping Deep Neural Network (DNN) $f : \mathbb{R}^D \rightarrow \mathbb{R}^d$ that
is shared across all nodes,
and an asymmetric edge function $g : (\mathbb{R}^d \times \mathbb{R}^d) \rightarrow \mathbb{R}$ to represent edges in the graph.
Our entire model $g(u, v) = f(Y_u)^T \times M \times f(Y_v)$ is end-to-end differentiable.
$M$ is low-rank, as $M=L \times R$, where both $L \in \mathbb{R}^{d \times b}$ and $R  \in \mathbb{R}^{b \times d}$ project the node manifold coordinates to smaller space $\mathbb{R}^b$.
Since $b$ is much smaller than $D$, we are able to reduce the final node embedding significantly.
Figure \ref{fig:model} shows a depiction of our architecture.
Our desired likelihood is quadratic but we estimate it with a tractable linear
objective using negative sampling, similar to \cite{word2vec}.

We find that explicitly modeling edges can drastically reduce the representation dimensionality, for both
directed and undirected graphs,  especially when coupled with a Deep Neural Network.
Further, modeling asymmetry by representing edge $(u, v)$ differently than $(v, u)$ gives an additional performance boost when preserving the structure of directed graphs.
We perform an extrinsic evaluation of our method, by comparing it to the state-of-the-art on link-prediction tasks over a
variety of graphs from social networks, biology, and e-commerce.
We show that we can consistently learn
orders of magnitude smaller embedding dimensions, while improving ROC-AUC metrics. For example, we reduce the error on directed graphs by up to $\approx 70\%$ and undirected graphs by up to $\approx 50\%$ when using same-sized representations.   However, when our model is restricted to representations which are \emph{8 times smaller} than the baselines, we reduce the error in some cases by up to $66\%$ on directed graphs and $16\%$ on undirected graphs.
We perform intrinsic evaluations, by training and rendering two-dimensional embedding spaces for two datasets,
which we use to gain intuition on placement choices made by our model.

To summarize, our contributions are as follows.
\begin{enumerate}
  \item We propose to explicitly model a directed edge function, which we define as low-rank affine
projections on a manifold that is produced by a Deep Neural Network (i.e. ``deep embeddings'').
  \item We propose a new objective function, the graph likelihood.
 \item These aspects significantly improve the state-of-art on learning continuous graph representations, especially on directed graphs, while producing significantly smaller representation spaces, as evaluated on five graph datasets.
\end{enumerate}

\section{Edge Representations}

It is common to embed a graph by learning one continuous $D$-dimensional vector $Y_u \in \mathbb{R}^D$ for every graph node $u \in V$, where
relationships between nodes $u$ and $v$ are captured in a very coarse way, through the use of a distance measure (e.g.\ $dist(Y_u,Y_v)$).
This \emph{node-centric} modeling assumes that all relationships in the graph are symmetric -- a limiting assumption which fails to capture any directed relationships. %

We seek to model the asymmetry which occurs in many real world graphs.
Specifically, given two nodes, $u$ and $v$, we desire that their distances are allowed to differ ($dist(Y_u,Y_v) \neq dist(Y_v,Y_u)$) to reflect ordering in directed relationships, such as \emph{follower} and \emph{followee} on Twitter. In addition, the asymmetry can also model degree variance in undirected graphs. Consider a popular node $m$, then the optimization could make $dist(Y_u, Y_m)$ small for all $u$ but not necessarily $dist(Y_m, Y_u)$. 

Even though it is possible to learn one representation $Y_{(u,v)}$ for all node pairs $(u,v)$, this direct modeling is prohibitive in practice and requires an upper-bound space of $O(|V|^2)$.
Instead, we propose to learn a trainable edge function defined over node embedding coordinates.
Specifically, we learn asymmetric transformations of the nodes, which generates for a node $u$, two representations: one when it is the source of a directed edge, $\hat{Y}^\text{source}_{u}$ and one when it is a destination, $\hat{Y}^\text{dest}_{u}$.
These representations share a neural network $f$.
These representations can be combined for any pair of nodes to model the strength of their directed relationships.
That is, for nodes $u$ and $v$, we represent $(u, v)$ and $(v, u)$ as
$dist(\hat{Y}^\text{source}_{u}, \hat{Y}^\text{dest}_{v})$ and $dist(\hat{Y}^\text{source}_{v}, \hat{Y}^\text{dest}_{u})$, respectively.

\section{Preliminaries}
\label{sec:preliminaries}

\subsection{Link Prediction}
Link prediction is a problem of inferring missing edges in a graph. We use it to evaluate the generalization ability of our embedding spaces, as we aim to preserve the graph structure. The common setup \cite{node2vec} is to ``hold out'' test edges $E_\textrm{test} \subset E$ and train on the remaining  $E_\textrm{train} = E - E_\textrm{test}$. Structure-preserving representations should retrieve the held-out $E_\textrm{test}$ with high accuracy.

\subsection{Graph Embedding}
Graph embedding approaches learn a $D$-dimensional embedding dictionary
${\bf Y} \in \mathbb{R}^{|V| \times D}$, containing continuous real-valued vector
$Y_u \in \mathbb{R}^D$ for every graph node $u \in V$. Earlier approaches in
computing embeddings include Eigenmaps \cite{eigenmaps}, which embeds $Y_u$ and $Y_v$ to be close if they are connected (i.e. $(u, v) \in E$ or similarly $A_{uv} =1$). Formally, Eigenmaps learns embeddings by minimizing an objective:
\begin{equation}ß
\label{eq:eigenmaps}
\min_{\textbf{Y}} \sum_{(u, v) \in E} A_{uv} ||Y_u - Y_v||^2_2 \quad  s.t. \quad \textbf{Y}^T \textbf{D} {\textbf{Y}} = I,
\end{equation}
where the weight of edge $(u, v)$ is stored in the adjacency matrix at $A_{uv}$ and ${\textbf{D}}$ is a
diagonal weight matrix with $D_{vv} = \sum_u A_{uv}$. This optimization yields an embedding space where
$Y_u$ and $Y_v$ are near if $A_{vu}$ is large (or non-zero).
Equation \eqref{eq:eigenmaps} also appears in equivalent forms in \cite{vlsi-spectral,shi-malik}. Furthermore, Bregman Iterations has been proposed to optimize an L1-formulation of the above objective function \cite{l1-segmentation}.

\begin{algorithm}[tb]
	\caption{Extract Random Walks}
	\label{alg:randomwalks}
	\begin{algorithmic}
		\STATE {\bfseries Input:} $G=(V, E)$, $n$ (walks per node), $\tau$ (walk length).
		\STATE {\bfseries Output:} \texttt{walks}.
		\STATE ${\bf \pi}$ \texttt{ = makeTransitionPr($E$)}
		\STATE \texttt{walks = []}
		\FOR{$u \in V$}
		\FOR{\texttt{(i = 0; i < $n$; i++)}}
		\STATE \texttt{walk = [$u$]};
		\FOR{\texttt{(j = 0; j < $\tau$; j++)}}
		\STATE \texttt{curr = walk[-1]}
		\STATE \texttt{$v$ = sampleNext(curr, ${\bf \pi}$)}
		\STATE \texttt{walk.append($v$)}
		\ENDFOR
		\STATE \texttt{walks.append(walk)}
		\ENDFOR
		\ENDFOR
	\end{algorithmic}
\end{algorithm}

\subsection{Word2vec}
Word2vec \cite{word2vec} processes a big text corpus (e.g. Wikipedia) and learns one embedding vector for every unique word. If two words $w_1$ and $w_2$ are frequently ``close'' (e.g. in same sentence), then the dot product of embeddings $Y_{w_1}^T Y_{w_2}$ is maximized. In particular, every time two words $w_1$ and $w_2$ are within $C$ words away, where integer $C$ is the ``context window size'' hyperparameter, then a gradient step increments this likelihood:
\begin{equation}
	\frac{\exp \left(Y_{w_1}^T Y_{w_2} \right)}{\sum_j \exp\left( Y_{w_j}^T Y_{w_2} \right) }.
\end{equation}
We refer the reader to \cite{word2vec} for further information\footnote{
	The denominator, rather than summing over all words, is approximated by hierarchical softmax. In addition, their original formulation learns two vectors per word, one when used as ``input'' and another used when ``output''.
}

\subsection{Graph Embedding with Random Walks}
Rather than operating directly on the adjacency matrix ${\bf A}$, another embedding strategy has been recently proposed by Perozzi, Al-Rfou, and Skiena \cite{deepwalk}.  
Their method, DeepWalk, introduced a new class of \textit{Random Walk} methods, which extend a node's direct neighbors to include nodes that are within small number of hops.
These approaches sample many random walks from the graph. If nodes $u$ and $v$ are frequently close in the random walks, then the model learns a representation such that the inner product of $\langle Y_u, Y_v \rangle$ is a large positive value.
Algorithm \ref{alg:randomwalks} extracts random walks. It begins by computing a probability transition matrix $\mathbf{\pi}$, where
$\pi_{u\rightarrow v}$ indicates the probability of a random walker visiting node $v$ conditioned on current node being $u$. 

This model has been extended by node2vec \cite{node2vec} to use a second-order probability transition function containing $\pi_{t \rightarrow u \rightarrow v}$, where the probability of a random walker visiting node $v$
is conditioned on current node $u$ and previous node $t$. 
Node2vec's random walk use hyper-parameters $p$ and $q$,
which effectively yield a graph traversal algorithm that's like an interpolation between Depth-First Search (DFS) and Breadth-First Search (BFS). We refer the reader to \cite{node2vec} for further details. We adopt this method for generating random walks in our work.

\begin{figure}[t!]
	\begin{center}
		\centerline{\includegraphics[width=\columnwidth]{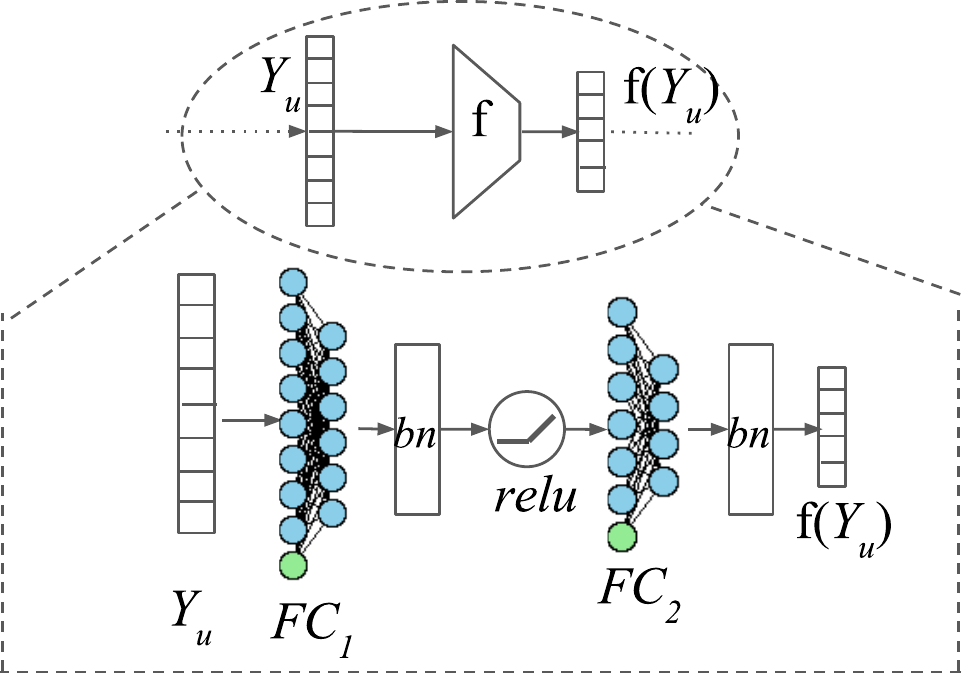}}
		\caption{Depiction of $f_\theta$, where $\text{FC}_i(x) = \mathbf{W}_i x + \mathbf{b}_i$ is a fully-connected layer with weight matrix $\mathbf{W}_i$ and bias vector $\mathbf{b}_i$, bn is BatchNorm \cite{batchnorm} and $\text{relu}(x) = \max(0, x)$ is an element-wise activation function.}
		\label{fig:dnn}
	\end{center}
	\vskip -0.1in
\end{figure}

After sampling random walks, DeepWalk and node2vec treat each walk
$(u_1 \rightarrow u_2 \rightarrow  \dots \rightarrow u_\tau)$
as a sequence, and then apply the skip-gram model
to compute embeddings per word (i.e. node).
The objective that they minimize is:
\begin{equation}
\label{eq:skipgram}
\min_\mathbf{Y} \left[ \log Z - \sum_{u \in V, v \in V} D_{uv} (Y_u^T Y_v) \right],
\end{equation}
Where $D_{uv}$ is the number of times nodes $u$ and $v$ appear close to each other (i.e. within the context size) in all random walks.
We extend these random walk methods in three important ways:
First, rather than using word2vec's objective (Equation \ref{eq:skipgram}), we propose an novel alternative objective, the \textbf{graph likelihood} (see Section \ref{sec:graphlikelihood}). Second, we explicitly represent an edge function as a function of nodes which we jointly train (see Section \ref{sec:model}). Third, we define the ``context'' for directly graph differently than undirected ones.
Specifically, a random walk $u_1\rightarrow u_2\rightarrow u_3\rightarrow u_4\rightarrow u_5$ would produce 
 $\{u_1, \dots, u_5\}$ as a context of $u_3$ if graph is
 undirected and would produce $\{u_4, u_5\}$ as context if the graph is directed.

\section{Our Method}
\label{sec:model}
We explain the details of our model and how we train it. The source-code is made available online \footnote{Code available at \url{http://sami.haija.org/graph/deep_embedding.html}} .

\subsection{Model}
\label{sec:g}
Given an (un)directed graph $G = (V, E)$, we learn an embedding vector $Y_u \in \mathbb{R}^D$ for
every node $u \in V$. In addition, we learn a Deep Neural
Network (DNN) $f_\theta : \mathbb{R}^D \rightarrow \mathbb{R}^d$
that maps a node onto a low-dimensional manifold. $f_\theta$ is depicted in Figure \ref{fig:dnn}, and is defined as:
\begin{align*}
f_\theta : Y_u            & \rightarrow \text{FC}_{\{\mathbf{W}_1, \mathbf{b}_1\}} \rightarrow \text{BatchNorm} \rightarrow \text{relu} \\
& \rightarrow \text{FC}_{\{\mathbf{W}_2, \mathbf{b}_2\}} \rightarrow \text{BatchNorm} \rightarrow f_\theta(Y_u),
\end{align*}
where $\text{FC}_{\{\mathbf{W}, \mathbf{b}\}}$ is a fully-connected layer with weight matrix $\mathbf{W}$ and
	bias vector $\mathbf{b}$, BatchNorm is described in \cite{batchnorm},
$\text{relu}(x) = \max(0, x)$ is an element-wise activation function,
and $\theta = \{\mathbf{W}_1, \mathbf{b}_1, \mathbf{W}_2, \mathbf{b}_2, \dots \}$.

We define a general class of edge functions $g(u, v) \in \mathbb{R}$ where
symmetricity is not imposed, yielding $g(u, v) \ne g(v, u)$. Consider a low-rank affine
projection in the manifold space:
        \begin{equation}
          \label{eq:g}
          g(u, v) = f(Y_u)^T \times M \times f(Y_v),
        \end{equation}
where low-rank projection matrix $M= L \times R$ with $L \in \mathbb{R}^{d \times b}$
and $R \in \mathbb{R}^{b \times d}$.
We refer to $b$ as the bottleneck dimension and we experiment with
$b < d < D$. We can factor $g(u,v)$ into an inner product
$\langle L^T f(Y_u), R f(Y_v) \rangle$. We refer to
$L^T f(Y_u) \in \mathbb{R}^b$ and $R f(Y_v) \in \mathbb{R}^b$, respectively, as the left- and right-asymmetric
embeddings.

We note Equation \eqref{eq:g}  can be extended to use a combination of multiple low-rank affine
projections, as:
\begin{equation}
g^{(2)}(u, v) = \langle w^{(2)}_g, [\text{relu}(g_1(u, v)), \dots, \text{relu}(g_h(u, v))] \rangle,
\end{equation}
where $w^{(2)}_g \in \mathbb{R}^h$ a parameter vector of the output layer, $h$ is the number of
projections, and each projection $g_i$ has its own $L_i \in \mathbb{R}^{d \times b}$
and $R_i \in \mathbb{R}^{b \times d}$. 
Even though there total size of parameters for ${\bf Y}, f, g$ may
be large, we can have a low memory footprint during inference if we precompute $L_i^T f(Y_u)$ and $R_i f(Y_u)$
for every $u \in V.$

\subsection{Graph Likelihood}
\label{sec:graphlikelihood}
We introduce our proposed objective, step-by-step. We start with intuitions from the Maximum Likelihood Estimate of Logistic Regression. Given a training graph $G = (V, E_\text{train})$, one can define a probability measure as a product of an edge estimate $Q$ on all node pairs:
\begin{equation}
\label{eq:edgeliklihood}
\Pr(G) = \prod_{(u, v) \in E_\text{train}} Q(u, v) \prod_{(u, v) \notin E_\text{train}} 1 - Q(u, v),
\end{equation}
where $Q : V \times V  \rightarrow [0, 1]$ is a trainable edge estimator. If $Q$ is a perfect estimator, then it should output $1$  on all $(u, v) \in E$ and should output $0$ on all $(u, v) \notin E$, which makes $Pr(x) = 1$ iff $x=G$. 
An equivalent form of equation \ref{eq:edgeliklihood} is:
\begin{equation}
\label{eq:edgeliklihood2}
\prod_{\pbox{20cm}{$u \in V$ \\ $v \in V$} } Q(u, v)^{\mathbbm{1}[(u,v) \in E_\text{train}]}  \left( 1 - Q(u, v) \right)^{\mathbbm{1}[(u,v) \notin E_\text{train}]}
\end{equation}
where indicator function $\mathbbm{1}[x] = 1$  if predicate $x$ is true and is $0$ otherwise.
Note that two product terms are mutually exclusive, as one of the powers $\mathbbm{1}[.]$ will evaluate to 1 and the other to 0.

Recent work shows that extending the neighbor-set of nodes beyond their direct connections via random walks, can improve generalization of prediction tasks such as link-prediction and node classification \cite{deepwalk, node2vec, tridnn}. Following this motivation, we propose to replace the binary edge presence $\mathbbm{1}[(u,v) \in E_\text{train}]$ in equation \eqref{eq:edgeliklihood2} by
simulated random walk statistics,
and formulate our proposed quadratic objective, the \textbf{graph likelihood} as:
\begin{equation}
\label{eq:likelihood}
\Pr(G) \propto \prod_{\pbox{20cm}{$u \in V$ \\ $v \in V$} } \sigma(g(u, v))^{\mathcal{D}_{uv}} (1 - \sigma(g(u, v)))^{\mathbbm{1}[(u, v) \notin E_\text{train}]}
\end{equation}
where $\sigma(x) = 1 / (1 + \exp(-x))$ is the standard logistic, the edge function $g$ is described in section \ref{sec:g}, and
$\mathcal{D}_{uv}$ is the unnormalized frequency that nodes $u$ and $v$
appear within the configured context window, in simulated random walks \cite{deepwalk}. Note that our likelihood in equation
\eqref{eq:likelihood} is not standard, especially that the two terms are not exclusive,
$0$ and $(u, v) \notin E_\text{train}$ to
be simultaneously true.  It follows that the expression under the product-operator is $\in [0, 1]$ since the logistic
$\sigma : \mathbb{R} \rightarrow [0, 1] $. We use \textit{proportional to} ($\propto$) instead of equality since the
normalizing constant $\int_G \Pr(G)$ is only a scaling factor and should not change the $\argmax$ of the likelihood.
Our experiments show that the likelihood yields a powerful representation for preserving the graph structure,
as evaluated on link-prediction tasks, out-performing models trained with a skipgram objective (such as node2vec \cite{node2vec}). We show in our Experiments (Section \ref{sec:experiments}) that even when our model is identical to node2vec (i.e. shallow and symmetric), training with our proposed objective produces embeddings that better preserve the graph structure, especially when using low embedding dimensions.

Although a na\"ive optimization of equation \eqref{eq:likelihood} is quadratic, $\mathcal{D}$ is sparse with $O(|V|)$ non-zero entries, making it possible to compute the first term in equation \eqref{eq:likelihood} in linear time.
Further, we use negative sampling (Section \ref{sec:nce}) to estimate the product over $(u, v) \notin E$,
which is important in many real applications where graphs are large and most edges are negative, having $|\{(u, v)  : u, v \in V \text{ and } (u, v)  \notin E  \}| \approx \mathcal{O}(|V|)$.

\subsection{Training Data Generation}
Our training algorithm requires positive and negative pairs of nodes as its input. Here we briefly describe their generation.

\subsubsection{Positives}
Given a graph $G = (V, E)$, we take a partition $E_\text{train} \subset E$ and
extract random walks from $E_\text{train}$ using Algorithm \ref{alg:randomwalks}.
Starting from every node $u_1$, we simulate $n$ random walks, each of length $\tau$, like:
\[
u_1 \rightarrow u_2 \rightarrow u_3 \rightarrow \dots \rightarrow u_\tau.
\]
Then, for every walk, we extract all node pairs within the context window, similar to \cite{word2vec}.
\begin{equation}
\label{eq:pairs}
(u_i, u_j) \quad \forall j \in \mathbb{Z}, i-w_l \le j \le i + w_r, j \ne i,
\end{equation}
where $w_l$ and $w_r$ are the context window left and right offsets. For example,
for an undirected graph, if we use a context window of size 5, then $w_l = w_r = 2$ and
therefore $j \in \{-2, -1, 1, 2\}$. Extracting positive pairs yields a list $\mathcal{D}$ that contain
duplicates and we over-load this notation by defining $\mathcal{D}_{uv}$ as the
frequency of $(u, v)$ in list $\mathcal{D}$. It is trivial to show that the
number of pairs is linear in $|V|$:
\begin{align}
\begin{split}
|\mathcal{D}| &= n \tau \mathcal{O}((w_l + w_r)(\tau - (w_l + w_r + 1)) |V|) \\
              &= \mathcal{O}(|V|)
\end{split}
\end{align}

\subsubsection{Negatives}
   We fix a set of negatives for every node. Before training, for every node
   $u$, we create its negative set $\bar{u}$ as:
   \begin{equation}
   \bar{u} = \{v_1^{-}, v_2^{-}, \dots \} \quad \text{s.t. } \quad \forall v^{-} \in \bar{u}, (u, v^{-}) \notin E_\text{train}
   \end{equation}
   where elements of $\bar{u}$ are sampled uniformly at random. Arguably, it is
   possible to increase the accuracy of our models if we sub-sample frequent nodes (i.e.\ with a high degree) as recommended by \cite{word2vec}, however we leave this
   as future work. In our training loop, we uniformly sample a subset of size
   $K$ from $\bar{u}$, where $K$ is a hyper-parameter for Negative Sampling. We use a fixed $K = 5$ for all our experiments.

\subsection{Negative Sampling}
\label{sec:nce}
We define an objective, $\mathcal{L}$, that can be computed in linear-time using negative sampling, (similar to \cite[Section 2.2]{word2vec}). $\mathcal{L}$  approximates our quadratic graph likelihood \eqref{eq:likelihood}, defined as:
\begin{align}
\label{eq:loglikelihood}
\begin{split}
\mathcal{L} = \mathop{\mathbb{E}}_{(u, v) \sim \mathcal{D}/Z} \bigg[ & \log \sigma(g(u, v))  \\
                              & + \sum_{v^{-} \in \text{Sample}(K, \bar{u})} \log(1 - \sigma(g(u, v^{-}))) \bigg],
\end{split}
\vspace{-0.1in}
\end{align}
where $\text{Sample}(K, \bar{u})$ uniformly samples $K$ negatives from $\bar{u}$
without replacement and $Z$ is a normalizing constant. Note that the outer expectation
$(u, v) \in \mathcal{D}$ is linear and the inner summation goes over $K$ items.
We use TensorFlow \cite{tensorflow} to obtain the gradients
$\frac{\partial \mathcal{L}}{\partial Y_u}, \frac{\partial \mathcal{L}}{\partial \theta},
\frac{\partial \mathcal{L}}{\partial L}, \frac{\partial \mathcal{L}}{\partial R}$
for each mini-batch. We use PercentDelta \cite{percentdelta} to optimize all parameters
$\theta, f, g, Y$. We only update the anchor embeddings $Y_u$
during the gradient steps on the objective $\mathcal{L}$, as
preliminary experiments showed that we get better performance. %

\section{Experiments}
\label{sec:experiments}

For all of our experiments, we simulated $n=80$ walks from every node, each walk is of length $\tau = 100$, and we used a right and left context window sizes, respectively, for directed and undirected graphs as $(w_l=0, w_r=2)$ and $(w_l=2, w_r=2)$.
\subsection{Datasets}

We test our algorithms on directed and undirected graphs. We obtain
PPI from \cite{ppi, node2vec} and the other datasets from Stanford SNAP \cite{snap}.
We only use the largest weakly connected component (WCC) from the original graph.
The statistics and dataset description are as follows:

\noindent \textbf{Directed graphs}:
\begin{enumerate}[topsep=0cm,labelsep=0cm,align=left,noitemsep,nolistsep]
	\item {\bf soc-epinions}: A social network $|V| = 75,877$ and $|E| = 508,836$. Each directed edge represents whether a user trusts the opinion of another.
	\item {\bf wiki-vote}: A voting network with $|V| = 7,066$ and $|E| = 103,663$. Nodes are Wikipedia editors. Each directed edges represents a vote that another becomes an administrator.
\end{enumerate}
\noindent \textbf{Undirected graphs}:
\begin{enumerate}[topsep=0cm,labelsep=0cm,align=left,noitemsep,nolistsep]
	\item {\bf ca-HepTh}: A citation network of High Energy Physics Theory from Arxiv, with $|V| = 17,903$ and $|E| = 197,031$. Each undirected edge represents co-authorship between two author nodes.
	\item {\bf ca-AstroPh}:  A citation network of Astrophysics from Arxiv, with $|V| = 17,903$ and $|E| = 197,031$. Each undirected edge represents co-authorship between two author nodes.
	\item {\bf PPI}: A protein-protein interaction graph, with $|V| = 3,852$ and $|E| = 20,881$. This is a challenging real-world dataset, where each node is a protein and an edge represents that two proteins interact.  
\item {\bf ego-Facebook}: A small portion of the Facebook social network, with $|V| = 4,039$ and $|E| = 88,234$. The nodes are users and the edges indicate friendship.
We note that this graph is an ego-network graph, which contains only the complete social connections of 10 seed users.  Rather than running link-prediction experiments on this graph, we analyze its unique structure through visualization in Section \ref{sec:visualization}.
\end{enumerate}

\subsection{Link Prediction}

We follow the setup in \cite{node2vec} for link prediction.
First, given a graph $G = (V, E)$, we partition its edges into two equal size
disjoint partitions $E_\text{train}$ and $E_\text{test}$, such that, $E_\text{train}$ is connected.
Second, we sample negative edges for training and testing, $E_\text{train}^{-}$
and $E_\text{test}^{-}$, where $E_\text{train}^{-}$ is sampled from the compliment of $E_\text{train}$ and
$E_\text{test}^{-}$ is sampled from the compliment of $E$. 
All train/test edge sets are of equal size.
Third, we simulate random walks on $E_\text{train}$ to get $\mathcal{D}$, using Algorithm \ref{alg:randomwalks} and Eq.\ \eqref{eq:pairs}.
Fourth, only for directed graphs, we extend $E_\text{test}^{-}$ to contain all edges
$(v, u)$ s.t. $(u, v) \in E$ and $(v, u) \notin E$. 
Finally, we train each algorithm and we evaluate ROC-AUC metrics on their ranking of
$(E_\text{test}, E_\text{test}^{-})$.

\begin{table*}[t!]
	\centering
	\setlength\tabcolsep{3.0pt}
	\begin{tabular}{ r@{\hskip 0.4cm} r || c | c | c || c | c | c | c | c | c | c | c | c }
& \textbf{Dataset} & \multicolumn{3}{c||}{\textbf{Adjacency Methods}} & \multicolumn{9}{c}{\textbf{Embedding Methods}} \\ 
\hhline{~-------------}
&     &   \multicolumn{3}{c||}{Non-Embedding baselines} & &  \multicolumn{3}{c|}{Embedding Baselines}    & \multicolumn{5}{c}{\textbf{Ours:} (end-to-end) Graph Likelihood} \\
& & \multirow{2}{*}{Jaccard}  & \multirow{2}{*}{\makecell{Common \\ Neighbors}} & \multirow{2}{*}{\makecell{Adamic \\ Adar}}  &     &   & &                 & \multicolumn{2}{c|}{Symmetric} & \multicolumn{2}{c|}{Asymmetric} & \multirow{2}{*}{\pbox{20cm}{ \% Error \\ Reduction}} \\
& &   &   &   & d & \makecell{Eigen \\ Maps} & node2vec & DNGR  &  shallow  &  deep  &  shallow  &  deep &   \\
\hhline{~=============}
\parbox[t]{2mm}{\multirow{10}{*}{\rotatebox[origin=c]{90}{directed}}\ldelim\{{10}{40pt}}
& \multirow{5}{*}{soc-epinions} & \multirow{5}{*}{0.649} & \multirow{5}{*}{0.649} & \multirow{5}{*}{0.647} & 8  & $\dagger$  & 0.725  & $\dagger$  & 0.694  & 0.665  & 0.695  & \textbf{0.825}  & 36.5\% \\ 
& & & & & 16  & $\dagger$  & 0.726  & $\dagger$  & 0.710  & 0.713  & 0.699  & \textbf{0.840}  & 41.4\% \\ 
& & & & & 32  & $\dagger$  & 0.714  & $\dagger$  & 0.740  & 0.713  & 0.700  & \textbf{0.845}  & 45.9\% \\ 
& & & & & 64  & $\dagger$  & 0.699  & $\dagger$  & 0.766  & 0.722  & 0.698  & \textbf{0.834}  & 44.9\% \\ 
& & & & & 128  & $\dagger$  & 0.691  & $\dagger$  & 0.782  & 0.743  & 0.718  & \textbf{0.828}  & 44.5\% \\ 
\cline{2-14}
& \multirow{5}{*}{wiki-vote} & \multirow{5}{*}{0.579} & \multirow{5}{*}{0.580} & \multirow{5}{*}{0.562} & 8  & 0.613  & 0.643  & 0.630  & 0.603  & 0.602  & 0.608  & \textbf{0.871}  & 63.7\% \\ 
& & & & & 16  & 0.607  & 0.642  & 0.622  & 0.623  & 0.639  & 0.643  & \textbf{0.900}  & 71.9\% \\ 
& & & & & 32  & 0.600  & 0.641  & 0.619  & 0.642  & 0.661  & 0.683  & \textbf{0.911}  & 75.2\% \\ 
& & & & & 64  & 0.613  & 0.642  & 0.598  & 0.660  & 0.672  & 0.702  & \textbf{0.917}  & 76.7\% \\ 
& & & & & 128  & 0.622  & 0.643  & 0.554  & 0.682  & 0.685  & 0.730  & \textbf{0.917}  & 76.8\% \\ 
\hhline{~=============}
\parbox[t]{2mm}{\multirow{15}{*}{\rotatebox[origin=c]{90}{undirected}}\ldelim\{{15}{40pt}}
& \multirow{5}{*}{ca-HepTh} & \multirow{5}{*}{0.765} & \multirow{5}{*}{0.765} & \multirow{5}{*}{0.765} & 8  & 0.786  & 0.731  & 0.706  & 0.855  & 0.848  & 0.605  & \textbf{0.879}  & 43.2\% \\ 
& & & & & 16  & 0.790  & 0.787  & 0.780  & 0.894  & 0.826  & 0.885  & \textbf{0.899}  & 51.9\% \\ 
& & & & & 32  & 0.795  & 0.858  & 0.829  & 0.896  & 0.886  & 0.884  & \textbf{0.911}  & 37.8\% \\ 
& & & & & 64  & 0.802  & 0.886  & 0.868  & 0.878  & 0.884  & 0.870  & \textbf{0.910}  & 21.3\% \\ 
& & & & & 128  & 0.812  & 0.901  & 0.897  & 0.891  & 0.897  & 0.820  & \textbf{0.916}  & 14.6\% \\ 
\cline{2-14}
& \multirow{5}{*}{ca-AstroPh} & \multirow{5}{*}{0.942} & \multirow{5}{*}{0.942} & \multirow{5}{*}{0.944} & 8  & 0.825  & 0.811  & 0.852  & 0.923  & \textbf{0.925}  & 0.592  & 0.917  & 44.1\% \\ 
& & & & & 16  & 0.825  & 0.833  & 0.877  & \textbf{0.950}  & 0.923  & 0.657  & 0.945  & 55.8\% \\ 
& & & & & 32  & 0.825  & 0.899  & 0.917  & 0.955  & 0.938  & 0.942  & \textbf{0.955}  & 46.1\% \\ 
& & & & & 64  & 0.824  & 0.934  & 0.939  & 0.948  & 0.936  & 0.936  & \textbf{0.958}  & 30.7\% \\ 
& & & & & 128  & 0.829  & 0.955  & \textbf{0.968}  & 0.953  & 0.936  & 0.939  & 0.957  & n/a \\ 
\cline{2-14}
& \multirow{5}{*}{PPI} & \multirow{5}{*}{0.766} & \multirow{5}{*}{0.776} & \multirow{5}{*}{0.779} & 8  & 0.710  & 0.733  & 0.583  & 0.746  & 0.763  & 0.550  & \textbf{0.804}  & 26.6\% \\ 
& & & & & 16  & 0.711  & 0.707  & 0.687  & 0.780  & 0.772  & 0.786  & \textbf{0.817}  & 36.7\% \\ 
& & & & & 32  & 0.709  & 0.691  & 0.741  & 0.779  & 0.784  & 0.794  & \textbf{0.833}  & 35.5\% \\ 
& & & & & 64  & 0.707  & 0.671  & 0.767  & 0.791  & 0.767  & 0.813  & \textbf{0.837}  & 30.0\% \\ 
& & & & & 128  & 0.737  & 0.698  & 0.769  & 0.795  & 0.787  & 0.799  & \textbf{0.841}  & 31.0\% \\ 
\end{tabular}

	\caption[Table caption text]{
		Link Prediction results from ranking $E_\text{test}$ across five graph datasets. 
		Numbers shown are the ROC-AUC.
		Row-wise: the first two datasets are directed and the last four are undirected graphs.
		Column-wise: The first three methods are adjacency (non-embedding) methods that use  the direct neighbors $N(u)$ and $N(v)$ for computing $g(u, v)$. We then list all embedding methods, preceded by d, the dimensionality of the learned embeddings. We report three embedding baselines followed by our four embedding methods.
		We compare embedding methods across different dimensionality $\{8, 16, 32, 64, 128\}$, marking in \textbf{bold} the top performer for the given dimensionality.
		For asymmetric embedding methods, we train with half of the dimensionality (= $\{4, 8, 16, 32, 64\}$) since in practice we need to store
		both sides of the embedding to compute an edge score, therefore every row contains the same total dimensions per node.
		The last column shows the relative reduction in error of our Asymmetric Deep model compared the best baseline. %
		 $\dagger$ indicates that the algorithm runs out of memory on our machine with 32 GB ram.
	}
	\label{table:auc}
	
\end{table*}

\subsubsection{Methods}
We report results from various methods, including non-embedding baselines, embedding baselines, and our proposed embedding methods.

\noindent \textbf{Adjacency (non-embedding) Baselines}:

These methods require $E_\text{train}$ during inference. Let $N(u)$ denote the list of node $u$'s direct neighbors, that are observed according to $E_\text{train}$. If the graph is directed, then $N(u)$ only stores outgoing edges. %
Adjacency baselines score an edge $(u, v)$ as a function of $N(u)$ and $N(v)$. We evaluate against the following:
\begin{enumerate}[topsep=0cm,labelsep=0cm,align=left,noitemsep,nolistsep]
	\item \textbf{Jaccard Coefficient} models the edge score as \[g(u, v)=\frac{|N(u) \cap N(v)|}{|N(u) \cup N(v)|}\]
	\item \textbf{Common Neighbors} models the edge score as \[g(u, v) = |N(u) \cap N(v)|\]
	\item \textbf{Adamic Adar}	models the edge score as \[g(u, v) = \sum_{x \in N(u) \cap N(v)} \frac{1}{\log(|N(x)|)} \]
\end{enumerate}

\noindent \textbf{Embedding Baselines}:

These methods use $E_\text{train}$ to learn embedding $Y_u$ for every graph node $u$. During inference, they use the learned embedding dictionary $\mathbf{Y}$ but \textit{not} the original graph edges. We compare against various state-of-the-art embedding methods.

\begin{enumerate}[topsep=0cm,labelsep=0cm,align=left,noitemsep,nolistsep]
 \item Laplacian \textbf{EigenMaps} \cite{eigenmaps} finds the lowest eigenvectors of the graph Laplacian matrix The eigendecomposition is real iff the Laplacian is symmetric. Therefore, we convert directed graphs to undirected ones during training. During inference, we define the edge scoring function as $g(u, v)=-||Y_u - Y_v||$.
  \item \textbf{node2vec} \cite{node2vec} learns embedding by simulating random walks on $E_\text{train}$ and minimizing the skipgram objective (Equation \ref{eq:skipgram}). We use the author's code to learn node embeddings, then we calculate the hadamard product $Y_u \odot Y_v$ for all node pairs $(u, v)$ in $E_\text{train}$ or $E_\text{train}^{-}$. Finally, we model an edge score as $g(u, v) = w^T(Y_u \odot Y_v)$ where $w$ is trained using off-the-shelve binary classification algorithm, scikit-learn's Logistic Regression. We train $w$ so that the logistic $\sigma(w^T(Y_u \odot Y_v)) \approx 1$ if $(u, v) \in E_\text{train}$ and $\approx 0$ if $(u, v) \in E_\text{train}^{-}$. According to our understanding, this is similar to how node2vec performed link prediction \cite{node2vec}.
  \item \textbf{DNGR} \cite{dngr} learns a non-linear (i.e. deep) node embeddings by passing ``smoothed'' adjancency matrix through a deep auto-encoder. The ``smoothing'' (called Random Surfing in \cite{dngr}) is their proposed alternative to random walks, which effectively has a different context weighing from node2vec. We use the author's code to train the auto-encoder on the adjacency matrix that corresponds to $E_\text{train}$. To test different embedding sizes, we only change the size of the last bottleneck layer in their code and keep the remainder of default architecture. We then output the bottleneck layer values for all nodes to and use them for the link prediction task, with scoring function $g(u, v) = Y_u^T Y_v$.
  \end{enumerate}
\noindent \textbf{Our Methods}:  
  \begin{enumerate}[topsep=0cm,labelsep=0cm,align=left,noitemsep,nolistsep]
  \item \textbf{Symmetric Shallow}: $w^T (Y_u \odot Y_v)$. From a modelling prospective, this is identical to node2vec's model. However, we train this model on our objective (Equation \ref{eq:loglikelihood}) rather than the skipgram objective (Equation \ref{eq:skipgram}).
  \item \textbf{Symmetric Deep}: $w^T (f(Y_u) \odot f(Y_v))$. Similar to above, except that the embedding representation is deep.  
  \item \textbf{Asymmetric Shallow}: $Y_u^T \times L \times R \times Y_v$. Applying asymmetry directly on the embeddings without a DNN.
  \item \textbf{Asymmetric Deep}: $f(Y_u)^T \times L \times R \times f(Y_v)$. Our full asymmetric formulation, when composed of a single affine projection. For both of our asymmetric methods, after training $\mathbf{Y}, f_\theta, g$, we use only the $b$-dimensional edge representations for inference ($L^T Y_u$ and $R Y_u$).
  
\end{enumerate}

We train all our models on  $\mathcal{D}$ using PercentDelta \cite{percentdelta}, with learning rate $0.001$ and L2 regularization of $0.0001$ on all trainable parameters. We find that PercentDelta is cruicial, especially that gradients w.r.t. embeddings were large with our initialization scheme. We do model selection using $E_\text{train}$ and $E_\text{train}^{-}$.

\begin{figure*}[ht]
\vskip 0.2in	
\begin{center}
\begin{subfigure}[b]{0.23 \textwidth}
	\includegraphics[width=1 \hsize]{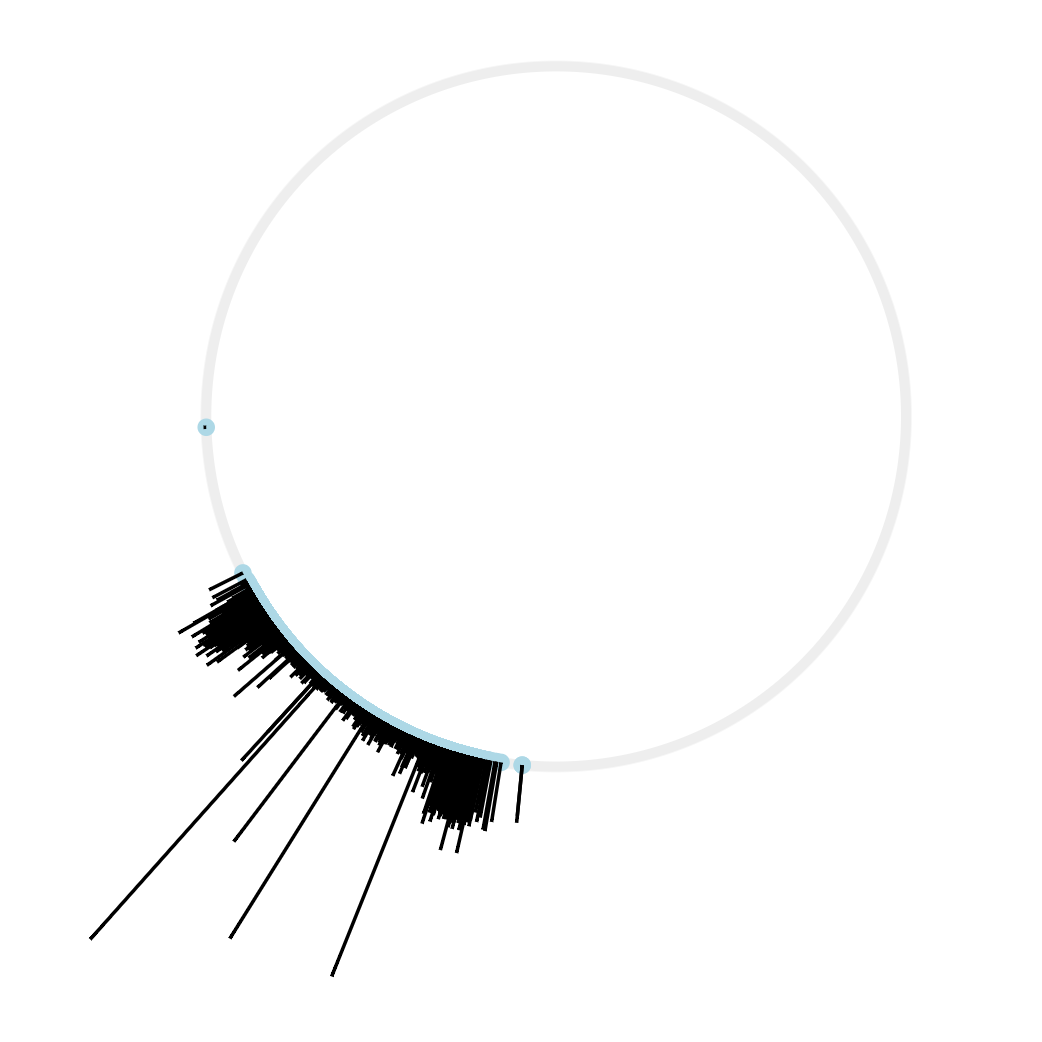}
	\subcaption{$L^T f({\bf Y})$ and degree}
\end{subfigure}
\begin{subfigure}[b]{0.23 \textwidth}
	\includegraphics[width=1 \hsize]{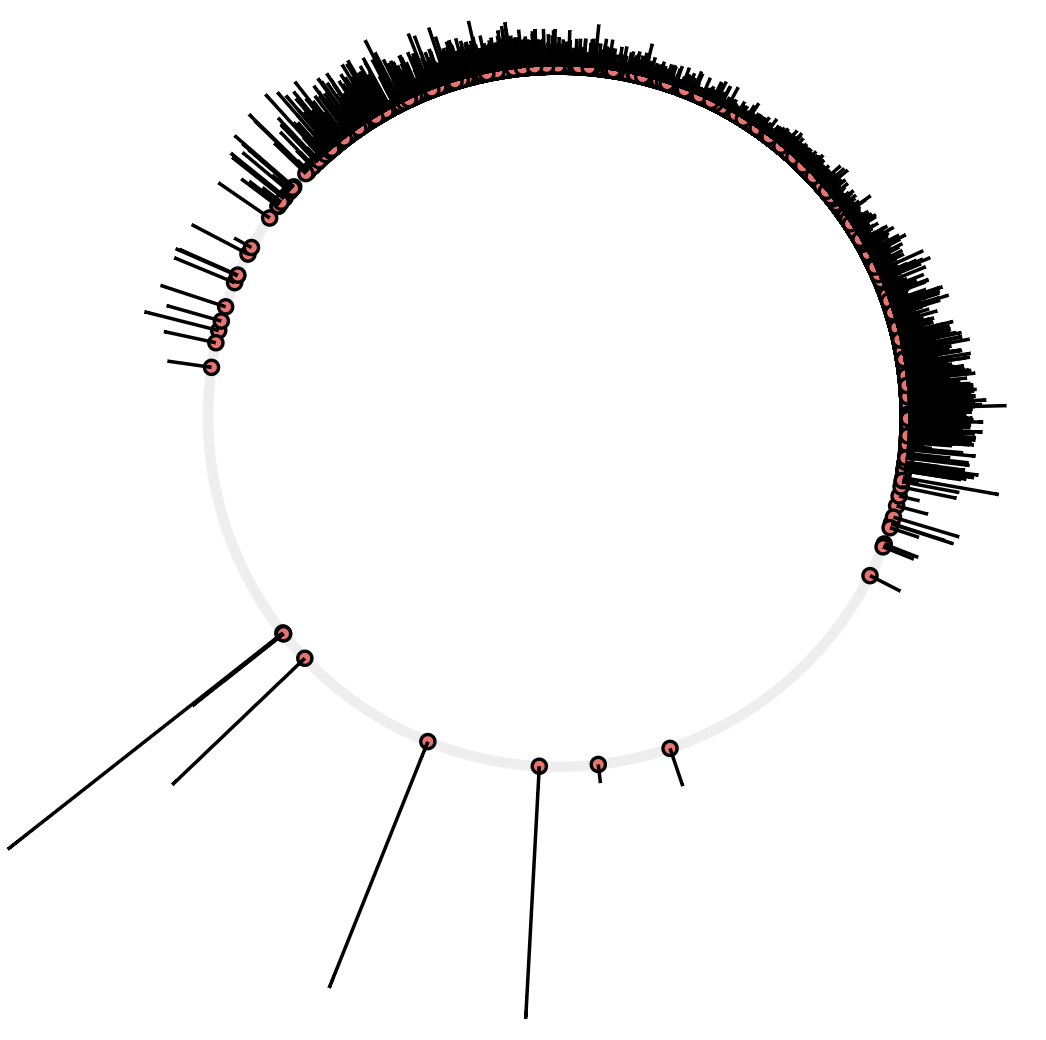}
	\subcaption{$R f({\bf Y})$ and degree}
\end{subfigure}
\begin{subfigure}[b]{0.23 \textwidth}
	\includegraphics[width=1 \hsize]{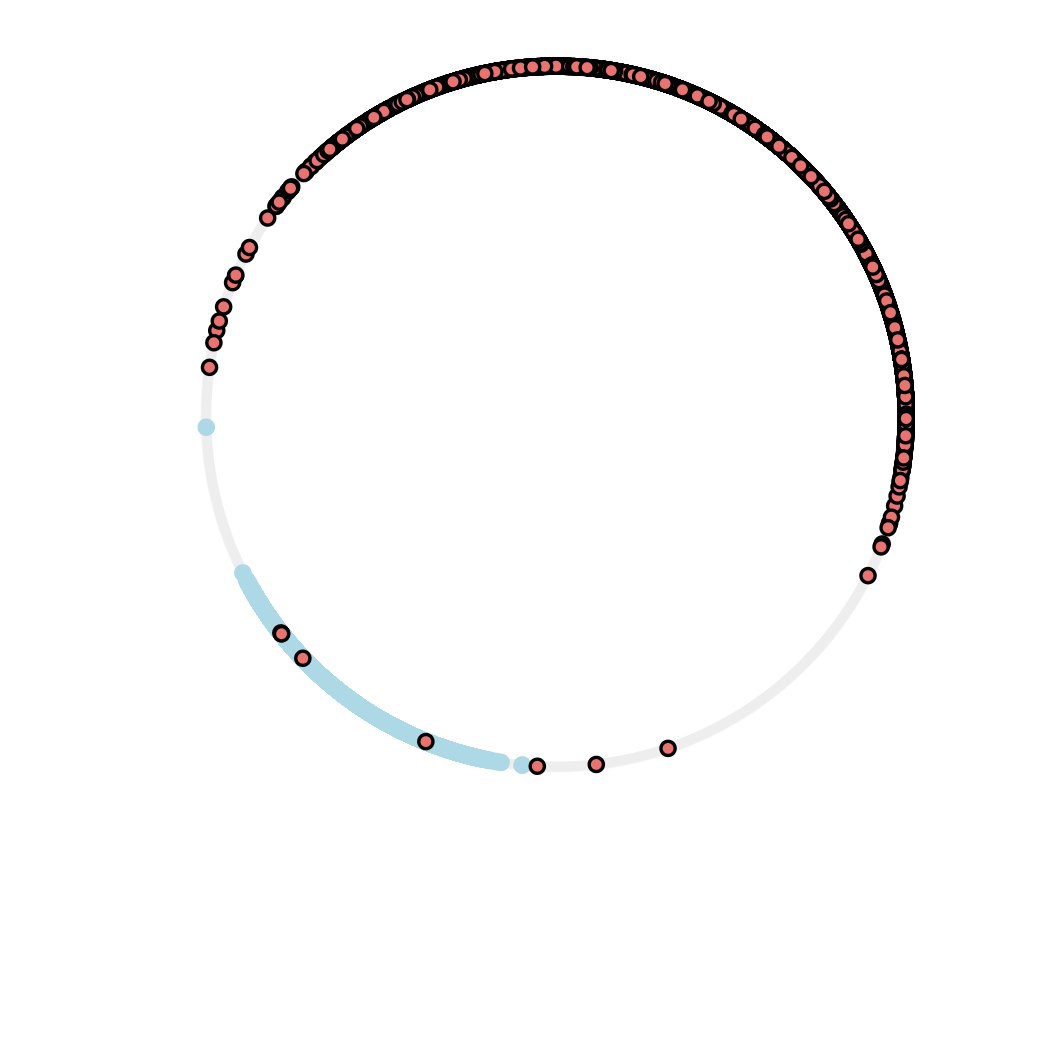}
	\subcaption{$L^T f({\bf Y})$ and $R f({\bf Y})$}
\end{subfigure}
\begin{subfigure}[b]{0.23 \textwidth}
	\includegraphics[width=1 \hsize]{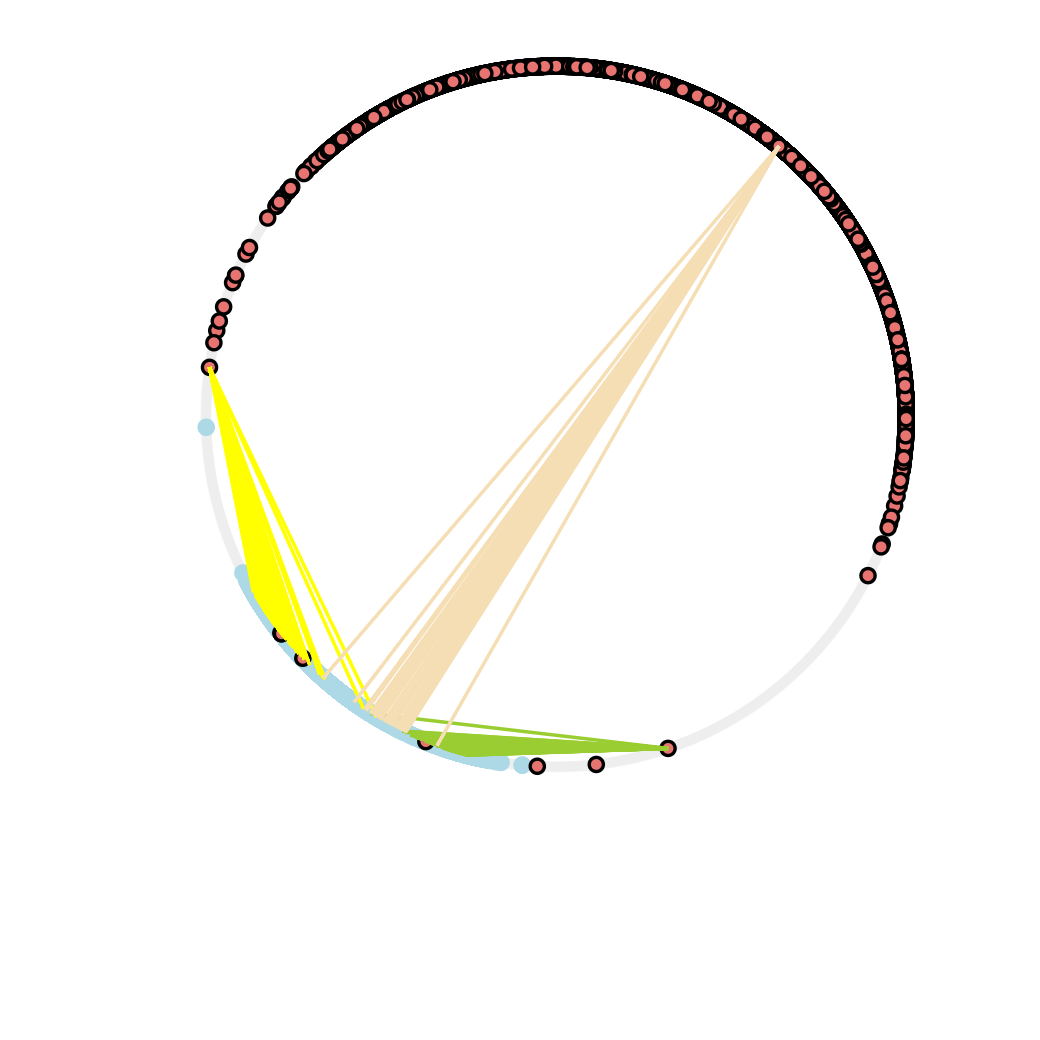}
	\subcaption{edges $R f({\bf Y}) \rightarrow L^T f({\bf Y})$}
\end{subfigure}
\caption{
Visualization of 2-dimensional embeddings learned from our Asymmetric Deep model on the Facebook dataset.
Figures (a-d) render nodes in their original learned coordinates (mapped to pixel space
as $(L^T \times f(Y_u)) * \text{radius} + \text{center}$).
Figs (a) and (b), respectively render the left- and right-embeddings of all nodes. 
Each node is plotted as a circle, with an overlay tick whose is length is proportional to its degree.
Fig (c) combines the left- and right-embedding spaces, dropping the degree ticks for clarity.
Fig (d) shows a few selected nodes from the right-embedding, and for each selected node, we draw all its
edges to nodes onto the left-embedding.
}
\label{fig:embeddings}
\end{center}
\end{figure*}

\begin{figure*}[ht]
\begin{center}
\begin{subfigure}[b]{0.24 \textwidth}
	\includegraphics[width=1 \hsize]{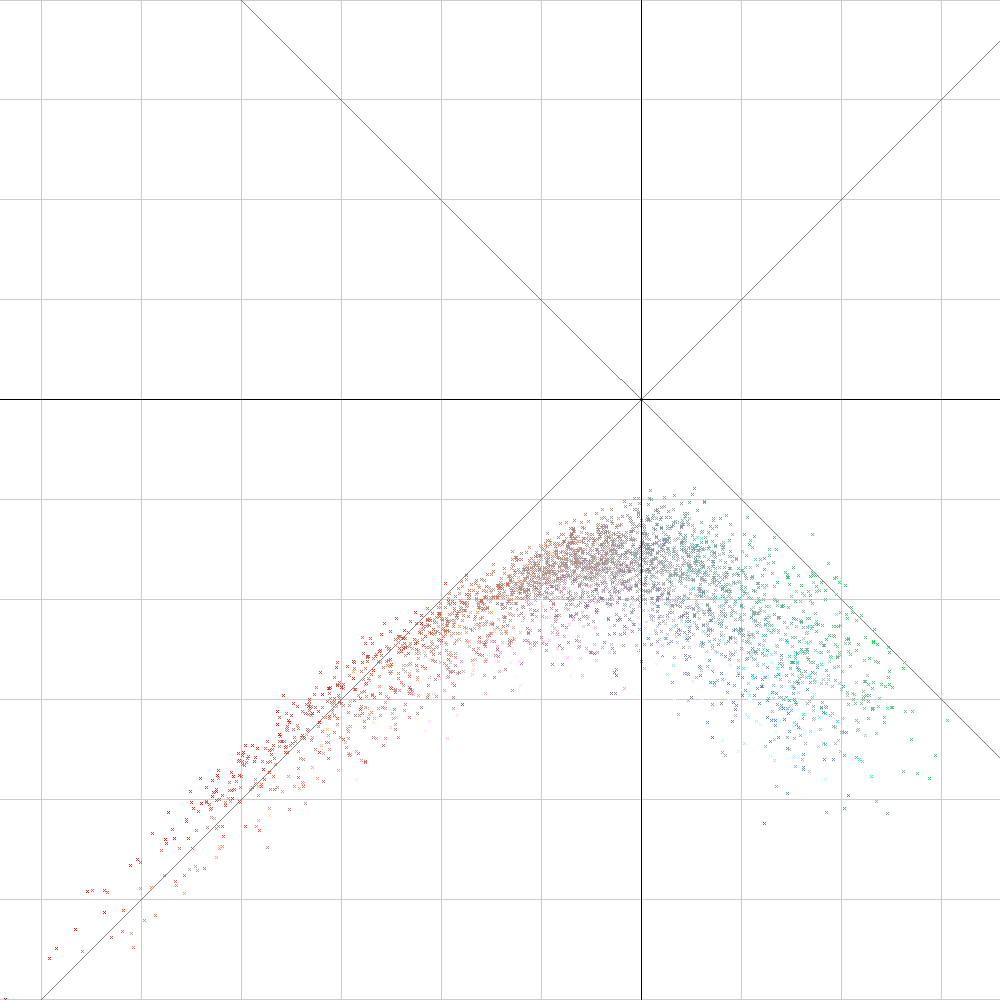}
	\subcaption{$L^T f({\bf Y})$}
\end{subfigure}
\begin{subfigure}[b]{0.24 \textwidth}
	\includegraphics[width=1 \hsize]{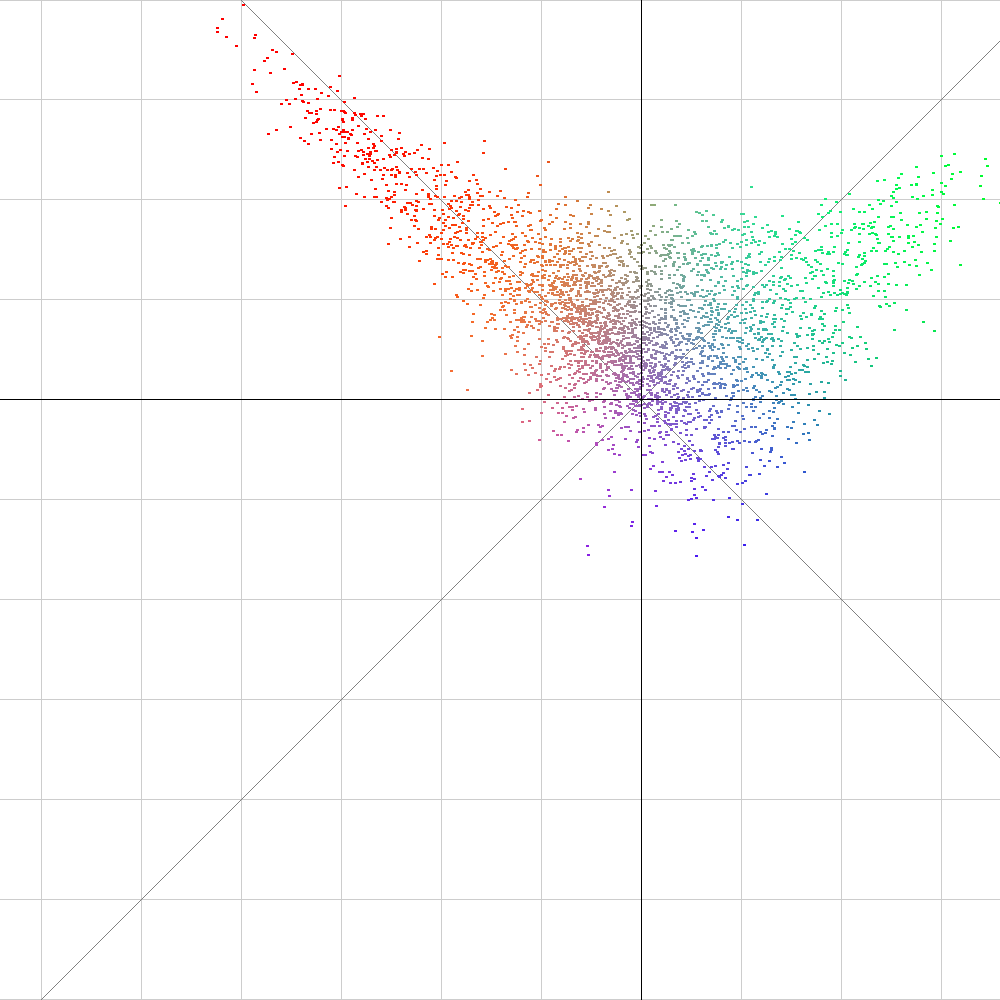}
	\subcaption{$R f({\bf Y})$}
\end{subfigure}
\begin{subfigure}[b]{0.24 \textwidth}
	\includegraphics[width=1 \hsize]{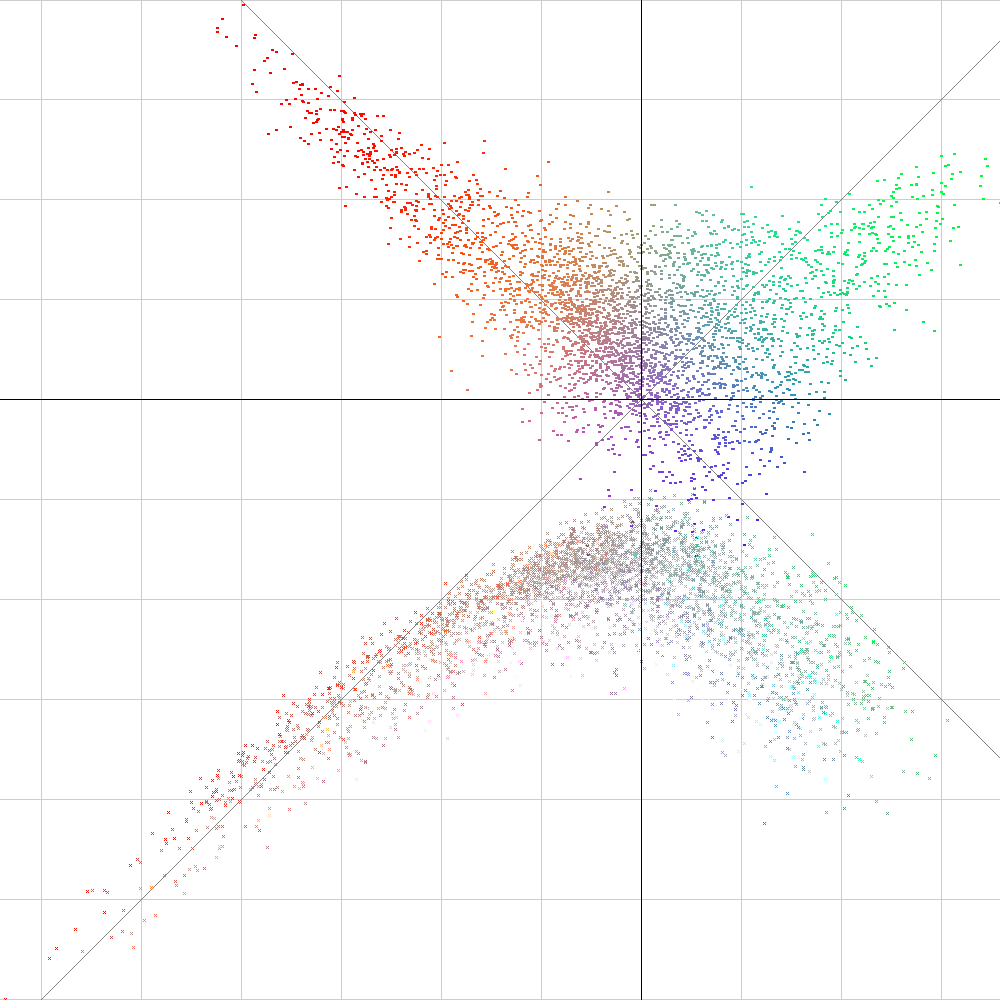}
	\subcaption{$L^T f({\bf Y})$ and $R f({\bf Y})$}
\end{subfigure}
\begin{subfigure}[b]{0.24 \textwidth}
	\includegraphics[width=1 \hsize]{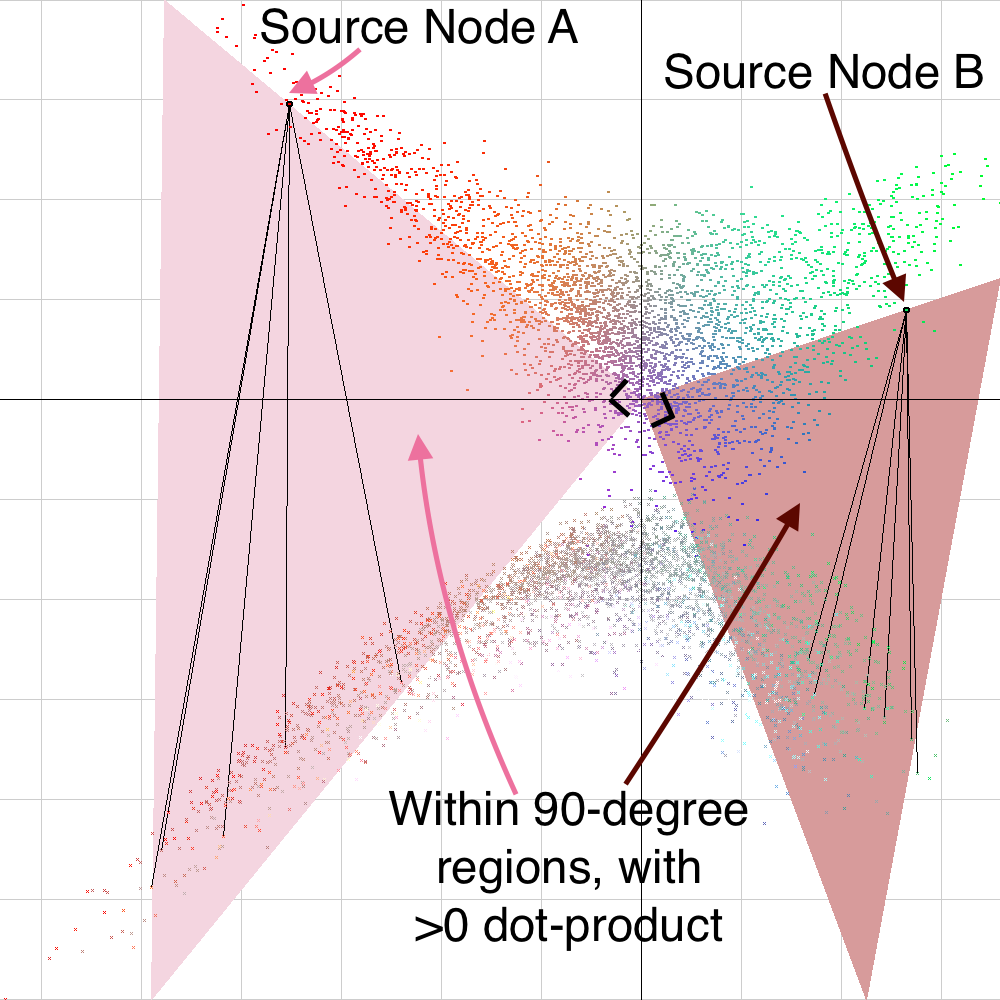}
	\subcaption{edges $R f({\bf Y}) \rightarrow L^T f({\bf Y})$}
\end{subfigure}
\caption{Unconstrained Asymmetric Embedding Visualization for PPI.  Figs (a) and (b) show the left- and right-embedding spaces.
Fig (c)  shows the combined plot.  Fig (d) shows two selected nodes from the right space and all of their edges onto the left space. 
The 90-degree cone per selected node, highlighting the area where the model has encoded probable edges (i.e.\ where the node has positive dot-product with the left embedding space).
The color of each node in the right embedding space depend on its (x, y) position. The colors of the left embedding space is set to the normalized adjacency matrix multiplied by the colors of the right embedding space.  
Axis lines $x=0$ and $y=0$ are shown in black, $y=x$ and $y = -x$ are shown in grey.}
\label{fig:ppiembeddings}
\end{center}
\end{figure*}

\subsubsection{Link Prediction Results}
Here we discuss the results of our link prediction experiments, which are presented in Table \ref{table:auc}.

First we turn our attention to directed graphs, where there is a dramatic increase in performance.
Specifically on graph soc-epinions, the asymmetric deep model reduces error over the baseline by 44.9\% for 64-dimensional representations.
We see that the asymmetric deep representations make more efficient use of their allocated space since its performance at lower dimensions (e.g. $d=8$) is approximately equal to its performance at higher dimensions (e.g. 64 or 128).
The second directed graph, wiki-vote, shows an even stronger performance increase, with a reduction in error of up to 76.8\% over the state-of-the-art baselines.

Next, we consider undirected graphs.  
On the citation networks, ca-HepTh and ca-AstroPh, we see that deep asymmetric approaches offer large improvements over the baseline when $d=16$ (respectively, 51.9\% and 55.8\%), but this lead narrows as the baseline representations are allowed more capacity.
Finally, we examine results on the protein-protein interaction network.
This is perhaps our most challenging undirected graph, derived from a problem of significant scientific interest.  
On this network, we observe a large boost from the deep asymmetric model over the baseline method, of up to a 36.7\% relative reduction in error ($d=64$).
We note that even when using representations which are \emph{16 times} smaller ($d=8$), the deep asymmetric model has AUC of 0.804, which is a 15\% error reduction over the best DNGR baseline ($d=128$, with AUC of 0.769).

We comment briefly on other observations from methods which optimize the graph likelihood.
First, from a modeling prospective, our shallow symmetric formulation is identical to node2vec's, but they differ in the training objective. This verifies that our proposed graph likelihood produces embeddings that better preserve the graph structure than the Skip-gram objective (Equation \ref{eq:skipgram}).
Second, shallow models tend to perform much worse in the presence of limited representation size.
This is unsurprising, as the a shallow model has to represent each node individually, rather than learning a common latent feature space which can impliclitly learn corelations in the data.
Third, using asymmetry alone (without a deep model) does not offer nearly as much performance improvement as the asymmetric deep model.

\begin{figure*}[t!]
	\begin{center}
		\includegraphics[width=\textwidth]{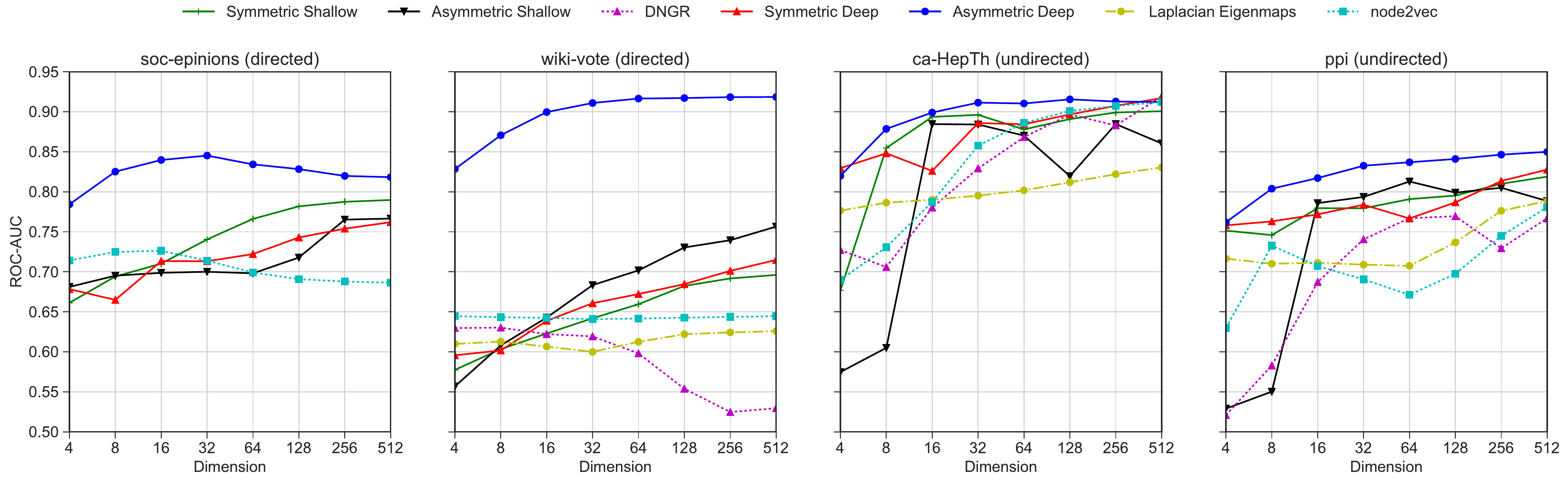}
		\vspace{-2.0em}
	\end{center}
	\vspace{-0.05in}		
	\caption{ROC-AUC results for our models versus the baselines measured on a suite of datasets.}
	\label{fig:auc-roc}
	\vspace{-0.05in}	
\end{figure*}

\subsection{Manifold Visualizations}
\label{sec:visualization}
We visualize asymmetric embeddings learned for link prediction for two graphs: 
ego-Facebook network and the
PPI network \cite{ppi}. We train both to be 2 dimensional ($b=2$), so that we can plot them on this paper without using an external embedding visualization algorithms such as t-SNE. To give two flavors of visualizations, we constraint the embedding of the former to be \textit{circular} but we put no constraints on the latter.

\subsubsection{Circular Visualization of ego-Facebook}
Similar to our link-prediction setup, we train on half of the edges (i.e. on $E_\text{train}$). However, we show on the visualization edges from both partitions $E_\text{train}$ and $E_\text{test}$.
In order to easily display the node degrees on the visualization, we constraint the embeddings to be circular (unit-norm) on both sides, specifically as:
$||L^T f(Y_u)||_2 = ||R f(Y_v)||_2 = 1$.
Figure \ref{fig:embeddings} shows embeddings. For and
setting L2-norm constraints on both sides of the asymmetric embeddings, specifically:
$||L^T f(Y_u)||_2 = ||R f(Y_v)||_2 = 1$.
Under this constraint, those 2-dimensional embeddings have only one degree of
freedom e.g. angle, and it is straight-forward to show that \[
\argmax_v \langle L^T f(Y_u), R f(Y_v) \rangle = \argmin_v || L^T f(Y_u) - R f(Y_v)||.
\] The visualization shows the two asymmetric embedding spaces
are almost disjoint,
where nodes from the right space are placed closer to their neighbors in the left space. We also note that the high-degree nodes are closest to the other embedding space. In fact, the highest degree nodes ``pull'' the left embedding space, as they live within it.

\subsubsection{Unconstrained Visualization of PPI}
We show in Figure \ref{fig:ppiembeddings} the left- and  right- embedding spaces learned for PPI when it is 2-dimensional ($b=2$). The right-embedding space was colored deterministically. In the right embedding space, the color of node $u$ is based on its right-embedding $(R f(Y_u)) \in \mathbb{R}^2$ coordinates. In the left embedding space, the color of node $v$ is set to the average of right-embedding colors of $v$'s neighbors. We see that nodes within 90 degrees have similar colors, showing that our method is embedding the nodes in appropriate positions across the two spaces, to preserve the graph structure.

\subsection{Improved Generalization}
Most machine learning models are prone to overfitting, showing higher performance metrics on the ``train'' partition than on the ``test'' partition. 
Here we consider an empirical evaluation of our proposed model's overfitting.
Specifically, we compute the ratio of test-over-train accuracy. If this test-over-train ratio $=1$, it means that the model does as well on (held-out) test data as it does on the training data. 
We note that this frequently does not occur in practice -- typically models overfit the training data, and the ratio is $<1$.
Nonetheless, Table \ref{table:generalization} shows that adding DNN $f()$ brings this ratio closer to 1. For example, on wiki-vote and soc-epinions, using the DNN $f()$ increases this ratio by over 4\%. 
Since we average this ratio across all our runs, we report the t-test numbers concluding that all our numbers are statistically significant ($p < 0.01 $).

In Table \ref{table:variance} we show that adding $f()$ can be seen as a regularization on the embeddings. In other words, deeper models produce $f(\mathbf{Y})$ with \textit{consistent} embedding L2-norms, across all datasets and across all runs, while shallow models produce $\textbf{Y}$ with a wider variation of L2-norms.

\begin{table*}[t!]
	\centering
	\begin{tabular}{r | c | c | c | c }
\multirow{2}{*}{Dataset} & \multicolumn{2}{c|}{mean$\left(\frac{\text{test AUC}}{\text{train AUC}}\right)$} & \multicolumn{2}{c}{Statistical Significance} \\
     & shallow asymmetric & deep asymmetric & t-statistic & p-value \\
\hline 
soc-epinions & $0.841$ & $0.882$ & 5.797673 & 1.53E-06\\ 
\hline 
wiki-vote & $0.908$ & $0.948$ & 3.881161 & 4.32E-04\\ 
\hline 
ca-HepTh & $0.881$ & $0.915$ & 5.202880 & 1.62E-05\\ 
\hline 
ca-AstroPh & $0.946$ & $0.970$ & 5.946066 & 4.08E-07\\ 
\hline 
ppi & $0.865$ & $0.893$ & 4.187474 & 8.45E-05\\ 
\end{tabular}

	\caption[Table caption text]{
		Showing generalization performance by averaging $\left(\frac{\text{test AUC}}{\text{train AUC}}\right)$ across all runs for all datasets under two settings: ``shallow'' VS ``deep'' asymmetric i.e. absence VS presence of DNN $f()$. Last two columns show the t-test for the difference of test-over-train AUC between the two settings.
		Adding a DNN to the model is a statistically significant improvement on all graphs with $p < 0.001$.  
	}
	\label{table:generalization}
\end{table*}

\begin{table*}[t!]
	\centering
	\begin{tabular}{r | c c c | c c c | c c c | c c c }
	Dataset & \multicolumn{3}{|c}{Shallow Symmetric} & \multicolumn{3}{|c}{Shallow Asymmetric} & \multicolumn{3}{|c}{Deep Symmetric} & \multicolumn{3}{|c}{Deep Asymmetric} \\
	& $25^\text{th}$ & $50^\text{th}$ & $75^\text{th}$  & $25^\text{th}$ & $50^\text{th}$ & $75^\text{th}$  & $25^\text{th}$ & $50^\text{th}$ & $75^\text{th}$  & $25^\text{th}$ & $50^\text{th}$ & $75^\text{th}$ \\
	\hline
	wiki-vote & 0.106 & 0.202 & 0.327 & 0.122 & 0.142 & 0.152 & 0.597 & 0.901 & 1.119 & 0.811 & 1.096 & 1.938 \\
	soc-epinions & 0.382 & 0.526 & 0.754 & 0.299 & 0.345 & 0.430 & 0.888 & 1.147 & 1.404 & 1.276 & 1.881 & 3.590 \\
	ppi & 1.884 & 4.593 & 7.842 & 0.858 & 1.197 & 3.801 & 0.825 & 1.095 & 1.410 & 1.015 & 1.443 & 2.645 \\
	ca-HepTh & 7.370 & 10.871 & 12.426 & 3.093 & 3.916 & 7.892 & 0.957 & 1.364 & 1.709 & 1.120 & 1.417 & 2.292 \\
	ca-AstroPh & 12.069 & 326.483 & 4282.095 & 1.108 & 4.648 & 24.271 & 0.874 & 1.273 & 1.999 & 1.065 & 1.826 & 3.062 \\
\end{tabular}
	\caption[Table caption text]{
		Inner-quartile Ranges of Standard Deviations of Embedding Norm, across all runs.
		For shallow models, we calculate $\text{std}_{u \in V}(||Y_u||)$, where $\text{std}_{u \in V}(.)$ is the standard deviation for all $u \in V$, and we display the statistics across all runs (e.g. different different embedding dimensions). For deep models, we calculate $\text{std}_{u \in V}(||f(Y_u)||)$
	}
	\label{table:variance}
\end{table*}

\subsection{Parameter Sensitivity}

In order to understand the impact of the representation size as a function of task performance, we varied the number of dimensions in the model from $d=4$ to $d=512$.
The results of this experiment are shown in Figure \ref{fig:auc-roc}.

\section{Discussion}
We have proposed a ``deep'' asymmetric model which learns $\mathbf{Y}$ jointly with $f$ and $g$ as $\mathbf{Y} \rightarrow f \rightarrow g$.
One might think that a ``shallow'' asymmetric model $\mathbf{Y} \rightarrow g$ should be able to learn $\mathbf{Y}$, identically to how a deeper counter-part learns $f(\mathbf{Y})$.
However this is not necessarily the case.
In this section we motivate why using the DNN $f()$ is helpful and reference empirical evidence that supports our claims.

First, $f()$ removes degrees of freedom as it passes an embedding $Y_u$ through  the DNN activation functions. Here, $f()$ can be seen as a regularizer over the embeddings. In particular, the output of $f(Y_u)$ is \textit{bounded}, as the last BatchNorm layer \cite{batchnorm} ensures that the $f(Y_u)$ has approximately a fixed mean and fixed variance across batches. On the other hand, shallow models can learn an \textit{unbounded} $Y_u$. We summarize the embedding norm statistics on real datasets in Table \ref{table:variance}.

Second, as $f()$ can constrain the embeddings with less degrees of freedom, this reduces overfitting and improves generalization. Table \ref{table:generalization} shows that deeper models have higher test-over-train accuracy metrics. This generalization is important since our proposed objective (the graph likelihood) is not directly a link prediction objective and the evaluation data $(E_\text{test}, E_\text{test}^{-})$ is not observed during training.

Finally, the hidden layers in $f()$ find correlations in the data, as many graph nodes have similar connections. 
It finds a smaller non-linear dimensional space that the nodes live in. For some $D$-dimensional embedding $\mathbf{Y} \in \mathbb{R}^{|V| \times D}$ neural network can map $\mathbf{Y}$ onto a lower dimensional manifold as $f : \mathbb{R}^D \rightarrow \mathbb{R}^d$ where $d < D$. In fact, our experiments in Table \ref{table:auc} show that our deep model performs quite well when provided with very few dimensions.

\section{Related Work}
\label{sec:related}
There is a rapidly growing body of literature on applying neural networks to 
problems which have as input a graph.
We divide the related work into two broad groups, based on whether
it concerns graph classification (e.g. assigning a label to $G$),
or learns a representation per node for preserving the graph structure.

\noindent \textbf{Discriminative Learning on Graph-Structured Data}.
These algorithms learn representations (at node/edge/graph) that are used for a discriminative classification task (per node/edge/graph). These methods are powerful for discrimination but they strictly rely on the graph structure as ``golden ground-truth'' to propagate information -- e.g. Graph-convolutional methods, using adjacency edges to define non-Euclidean patches \cite{patchysan, diffusion-cnn, bruna13} and some operate in the fourier domain \cite{bruna13, cnn-graph-struct, fast-spectrals}. In addition, some discriminative representations include fixed-point methods, recursively defining node features as a function of its neighbors by "unrolling a few steps" \cite{conv-graph-fingerprint} or until fixed-point convergence is reached \cite{gnn, gnn2, ggsnn, variational-struct}. We differ from all these methods, since they receive an external loss (e.g. label) and assume that the graph is completely observed. For example conditional independence assumptions made by the Markov Models of \cite{variational-struct} explicitly use the graph structure. Unlike our work, these methods have no obvious way
to estimate the score/probability of an edge, as the existence of the edge was inhertly used to pass discriminative information.

\noindent \textbf{Structure-Preserving Embeddings}. 
These methods learn one embedding per graph node, with an objective that maximizes (/ minimizes) the product (/ distance) of node embeddings if they are neighbors in the input graph. They are most related to our work.
In fact, our work builds on the approach introduced by Deepwalk \cite{deepwalk}, which 
learns node embeddings using simulated random walks.
These node embeddings have been used as features for various tasks 
on networks, such as node classification~\cite{deepwalk}, 
user profiling~\cite{Perozzi:2015:EAP:2740908.2742765}, and link prediction~\cite{node2vec}.
Some extensions of Deepwalk include: 
Walklets~\cite{walklets}, skipping nodes in the random walk to discover hierarchical structure;
node2vec~\cite{node2vec},
parameterizing the random walk process to allow more focused discovery of
structural relationships; author2vec~\cite{author2vec}, augmenting nodes with bag-of-word
representations for documents; and Tri-Party DNN~\cite{tridnn}, modeling heterogeneous graphs with
three different node types. 
Other node-centric methods are concerned with shorter dependencies in the graph~\cite{deep-struct-embedding,rdl}.
Finally, meta-embedding approaches, such as  HARP~\cite{chen2017harp}, have been proposed as general methods for improving node representations.

Our work differs from existing random walk methods in three ways.
First, we explicitly model asymmetric relationships between nodes. Even though random
walk methods we surveyed {\bf respect} edge direction during the walk, they do {\bf not} model
edge direction and represent $(u, v)$ identically to $(v, u)$ \cite{deepwalk, kgembed, author2vec, walklets, node2vec}.
This flexibility better models the heterogeneity which occurs in real world networks,
where social relationships may not be reciprocal (i.e.\ the graph is \emph{directed}), 
or typically have a very unbalanced degree distribution.
Second, our node embeddings are produced by a deep neural network (DNN), unlike the shallow (effectively 
1-layer) networks previously used. Third, rather than a 2-stage optimization of first training embeddings
on random walk sequences, followed by learning a task-specific classifier, we propose a graph likelihood and use it
to jointly train the embeddings, the manifold-mapping DNN, and the edge function. Even though
the graph likelihood does {\it not} match a link-prediction loss, it produces superior results on link-prediction
using fewer dimensions.

\section{Conclusion}
\label{sec:conclusion}
We introduced a novel method for integrating directed edge information for learning continuous representation for graphs.
Our method \emph{explicitly models edges} as functions of node representations.
We optimize this model using a new objective function, the \emph{graph likelihood}, which we
use to jointly learn the edge function and node representations.

Our empirical evaluation focused on link prediction tasks using a number of graphs collected from real world applications.
Our experimental results show that our proposed objective is better than the skipgram objective even when the model are identical. Our results also show that modeling edges as asymmetric affine projections through the node representation space,
helps produce more accurate and compact embedding spaces.
In particular, we show that asymmetric edge modeling, when trained with our objective, improves performance over state-of-the-art approaches,
especially on directed graphs, reducing error by up to $\approx 70\%$ and  $\approx 50\%$, respectively, on directed and undirected graphs while using the same number of dimensions per node.
We create visualizations to interpret the learned representations, showing that our method embeds nodes in appropriate places along the embedding manifold.

In addition to AUC metric improvements, explicit edge modeling allows us to learn smaller embeddings.
The representations learned through our model are more efficient at utilizing the available space. Our embeddings are able to outperform the baseline even
when outputting 8x fewer dimensions per node.
We believe that explicitly modeling edge representations addresses a substantial problem in the related work,
and can enable many avenues of future investigation for learning continuous representation of graphs.

\bibliographystyle{ACM-Reference-Format}
\bibliography{edge_learning} 

\end{document}